\documentclass[11pt,a4paper]{article}
\usepackage[T1]{fontenc}
\usepackage[utf8]{inputenc}
\usepackage[margin=25mm]{geometry}
\usepackage{lmodern}
\usepackage{amsmath,amssymb,graphicx,booktabs,tabularx,array,xcolor}
\usepackage{multirow}
\usepackage[super,sort&compress,numbers]{natbib}

\usepackage{url}
\let\nolinkurl\url
\usepackage{xurl}
\usepackage{microtype}
\usepackage{caption}
\usepackage{float}
\usepackage{needspace}
\usepackage[section]{placeins}
\usepackage{flafter}
\usepackage[hidelinks,bookmarksopen=true,bookmarksnumbered=true,pdfstartview=FitH]{hyperref}
\usepackage{bookmark}
\hypersetup{pdfpagemode=UseOutlines,pdftitle={A 10K drone image dataset with component-level annotations for rooftop photovoltaic inspection}}
\makeatletter
\def\hyper@link#1#2#3{#3\Hy@xspace@end}
\let\hyper@linkstart\@gobbletwo
\def\hyper@linkend{\Hy@xspace@end}
\makeatother
\newcommand{\topic}[1]{%
  \par\addvspace{0.6\baselineskip}%
  \Needspace{3\baselineskip}%
  \phantomsection\addcontentsline{toc}{subsection}{#1}%
  \noindent\textbf{#1.}\enspace\ignorespaces}
\renewcommand{\arraystretch}{1.18}


\newcommand{\optionalfigure}[3]{\includegraphics[width=#2,height=0.70\textheight,keepaspectratio]{Scientific_Data_assets/#1}}
\begin{document}
\title{A 10K drone image dataset with component-level annotations for rooftop photovoltaic inspection}
\author{\fontsize{11}{13.5}\selectfont Penglei Zhang$^{1,2}$, Wan Zhang$^{1,2,*}$, Ruizhe Deng$^{1,2,3}$, Bo Chen$^{4}$, Wenhao Duan$^{4}$,\\[0.3em]
\fontsize{11}{13.5}\selectfont Jinhe Zhang$^{4}$, Yanzhen Zhang$^{5}$, Qi Chen$^{4}$, Zhiling Guo$^{3}$, Haoran Zhang$^{5}$, Yuntian Chen$^{1,2}$}
\date{}
\maketitle
\begin{center}\small
$^{1}$Ningbo Institute of Digital Twin,
Eastern Institute of Technology, Ningbo, Zhejiang 315200, P.R. China\\
$^{2}$Zhejiang Key Laboratory of Industrial Intelligence and Digital Twin,
Eastern Institute of Technology, Ningbo, Zhejiang 315200, P.R. China\\
$^{3}$Department of Building Environment and Energy Engineering,
The Hong Kong Polytechnic University, Kowloon, Hong Kong, China\\
$^{4}$School of Geography and Information Engineering,
China University of Geosciences (Wuhan), Wuhan 430074, China\\
$^{5}$School of Urban Planning and Design,
Peking University, Shenzhen, 518055, China
\end{center}
\noindent$^{*}$Corresponding author: Wan Zhang; email: \nolinkurl{wzhang@idt.eitech.edu.cn}.

\pdfbookmark[1]{Abstract}{abstract}
\begin{abstract}
Component-level information supports rooftop photovoltaic (PV) inspection, but recognizing individual components across complex scenes remains challenging.
We present PVIS-10K, a dataset of 10,065 drone images with component-level annotations covering 63,632 physical PV modules across 22 rooftop installations.
The high-resolution red-green-blue (RGB) images were acquired by unmanned aerial vehicles (UAVs, also known as drones) in Ningbo, Shaoxing and Shenzhen, China.
Multi-angle oblique observations capture installation-height and scale variation, dense arrangements, occlusion, shadows and reflections.
Three-dimensional delineation, reprojection and image-by-image review produced 3,516,675 image-level polygon annotations, stored in Common Objects in Context (COCO) format.
Station-disjoint partitions support evaluation on unseen stations within and across cities.
Four supervised models and text-prompted zero-shot Segment Anything Model~3 (SAM~3) provide full-resolution benchmarks of instance matching and foreground coverage.
PVIS-10K supports component localization, delineation and image-based counting for drone-based rooftop PV inspection research.
\end{abstract}

\section{Background \& Summary}\label{sec:background}

Solar photovoltaic (PV) deployment continues to expand, with distributed rooftop installations forming an important part of this growth\cite{pvps2026}. The growing installed asset base increases the need for inspection and maintenance to sustain system performance\cite{pvpsom2022}. Component locations, boundaries and arrangement provide a spatial reference for these activities and support image-based counting.

Unmanned aerial vehicles (UAVs, also known as drones) provide high-resolution rooftop observations from different viewing angles\cite{colomina2014}, supporting component mapping for inspection\cite{r4}.
PV inventories establish facility locations\cite{r1,r2,r3,r13}, while aerial annotations and detection and segmentation studies map array boundaries and PV coverage\cite{r7,r11,r12,r8,r9,r10}.
For component-level inspection, these spatial extents must be complemented by boundaries that distinguish neighbouring physical components.

\begin{figure}[!htbp]\centering
\optionalfigure{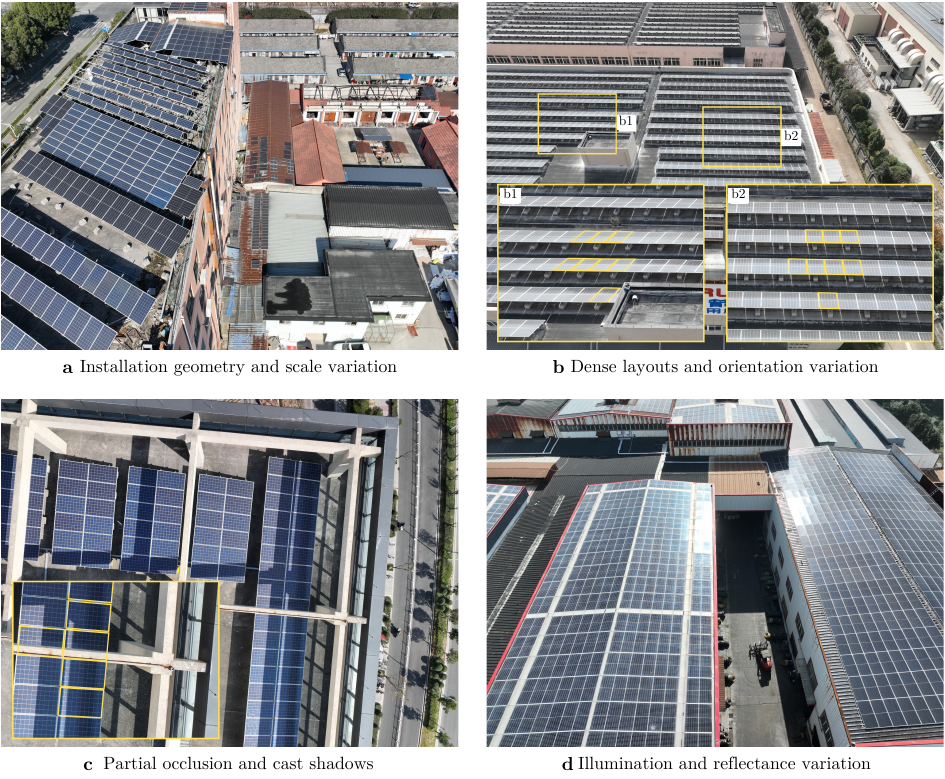}{\linewidth}{Representative rooftop PV scenes}
\caption{Challenges in drone-based rooftop PV component recognition.
Panels \textbf{a}--\textbf{d} show installation geometry and scale variation, dense layouts, partial occlusion and shadows, and illumination and reflectance variation, respectively.
Insets b1 and b2 in \textbf{b} and the lower-left inset in \textbf{c} show enlarged regions.
Yellow outlines indicate selected ground-truth instances, including projected occluded portions.}\label{fig:scene}
\end{figure}

Distinguishing components in rooftop imagery is complicated by scene geometry and viewing conditions (Fig.~\ref{fig:scene}).
Installation height, viewing angle and distance affect projected component sizes and shapes (Fig.~\ref{fig:scene}a), while dense layouts and repetitive textures complicate component separation (Fig.~\ref{fig:scene}b).
Partial occlusion, shadows and reflections further obscure boundaries or alter appearance (Fig.~\ref{fig:scene}c,d).
Internal dividing lines can also resemble boundaries between components.
Treating neighbouring components as one instance, or splitting one component, changes image-level counts even when the overall PV foreground remains similar.

\FloatBarrier

\Needspace{3\baselineskip}
The open-access red-green-blue (RGB) datasets compared in Table~\ref{tab:dataset_comparison} describe PV arrays or coverage regions.
Bradbury, BDAPPV and Queens provide aerial array annotations\cite{r7,r11,r12}; PV01 and Emmen support semantic segmentation from orthographic imagery\cite{r9,luimstra2025}.
These resources identify PV extent, but their labels do not separate neighbouring physical components.
Their orthographic or overhead representations also differ from multi-angle rooftop observations.
Complementary data are needed to evaluate component separation across changing viewing angles, projected scales and occlusion conditions.

\FloatBarrier
\begin{table}[!htbp]\centering\footnotesize
\caption{Representative open-access datasets for rooftop PV mapping.}\label{tab:dataset_comparison}
\begingroup
\setlength{\tabcolsep}{2pt}
\renewcommand{\arraystretch}{1.12}
\renewcommand{\tabularxcolumn}[1]{m{#1}}
\setlength{\aboverulesep}{2pt}
\setlength{\belowrulesep}{2pt}
\begin{tabularx}{\linewidth}{@{}>{\raggedright\arraybackslash}m{0.115\linewidth}>{\centering\arraybackslash}m{0.085\linewidth}>{\centering\arraybackslash}m{0.105\linewidth}>{\centering\arraybackslash}m{0.09\linewidth}>{\raggedright\arraybackslash}m{0.10\linewidth}>{\raggedright\arraybackslash}m{0.125\linewidth}>{\raggedright\arraybackslash}m{0.16\linewidth}>{\raggedright\arraybackslash}X@{}}\toprule[0.6pt]
Dataset & \shortstack[c]{Image\\count} & \shortstack[c]{Annotation\\count} & GSD & \shortstack[l]{Target\\level} & \shortstack[l]{Annotation\\type} & \shortstack[l]{Annotation\\workflow} & \shortstack[l]{View\\representation}\\\midrule[0.4pt]
Bradbury\cite{r7,bradburydata} & 601 & 19,433 & 30 & Non-\newline component & Semantic\newline segmentation & Manual 2D annotation & Orthographic view\\
\midrule[0.3pt]
BDAPPV\cite{r11} & 46,132 & \textemdash & 10; 20 & Non-\newline component & Semantic\newline segmentation & Crowdsourced 2D annotation & Overhead views\\
\midrule[0.3pt]
Queens\cite{r12} & 667 & $\sim$14,000 & 15.2 & Non-\newline component & Semantic\newline segmentation & Manual 2D annotation & Orthographic view\\
\midrule[0.3pt]
PV01\cite{r9} & 645 & 5,252 & 10 & Non-\newline component & Semantic\newline segmentation & \textemdash & Orthographic view\\
\midrule[0.3pt]
Emmen\cite{luimstra2025,emmendata} & 5,327 & 4,389 & 7.5 & Non-\newline component & Semantic\newline segmentation & Manual 2D annotation & Orthographic view\\
\midrule[0.3pt]
\textbf{PVIS-10K}\newline\textbf{(ours)} & 10,065 & 3,516,675 & $\sim$0.5--1.1 & Component & Instance\newline segmentation & 3D annotation and 2D reprojection & Multi-angle oblique views\\\bottomrule[0.6pt]
\end{tabularx}
\endgroup
\par\smallskip\raggedright Component and non-component denote individual components and arrays or coverage regions, respectively.
GSD, ground sampling distance (cm/pixel); \textemdash, not reported or undetermined.
BDAPPV provides 20,989 image masks rather than a reported instance total.
PVIS-10K GSD is a nominal nadir-equivalent estimate.
\end{table}
\FloatBarrier

\Needspace{3\baselineskip}
Delineating and checking individual component boundaries is labour-intensive, particularly when the same components appear in multiple views.
We developed PVIS-10K through three-dimensional annotation, two-dimensional reprojection and image-by-image correction.
It contains 10,065 multi-angle UAV images from 22 rooftop PV stations in Ningbo, Shaoxing and Shenzhen.
Its 3,516,675 image-level polygon annotations represent 63,632 physical components and are stored in Common Objects in Context (COCO) format\cite{r16}.
The dataset combines component-level instance boundaries with multi-angle oblique imagery of dense layouts, height and scale variation, occlusion, shadows and reflections.
Station-disjoint partitions support evaluation on unseen stations within and across cities, using supervised and text-prompted zero-shot benchmarks of component matching and foreground coverage.
These resources support component localization, delineation and image-based counting for drone-based rooftop PV inspection research.

\FloatBarrier
\section{Methods}\label{sec:methods}
\begin{figure}[!htbp]\centering
\optionalfigure{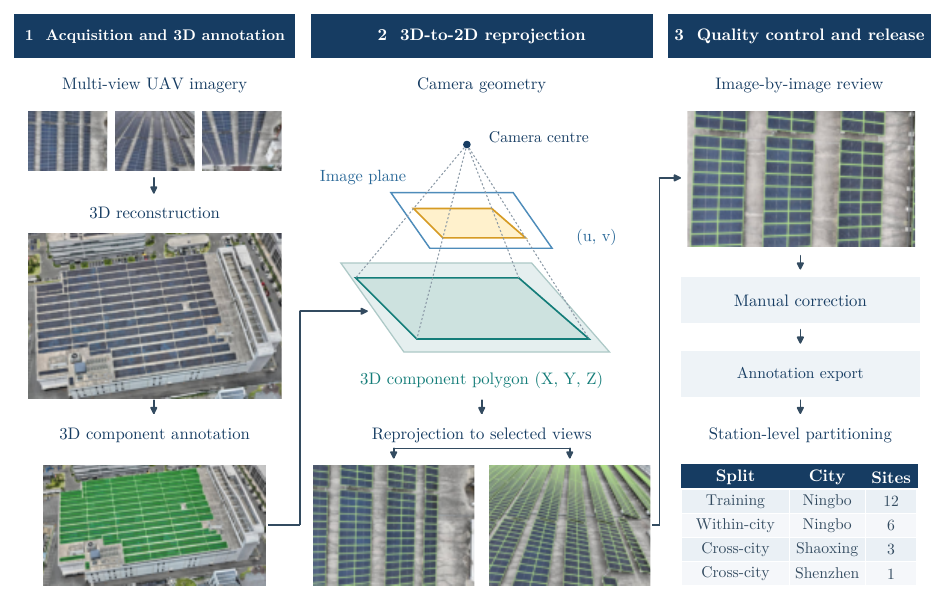}{\linewidth}{Dataset construction and release workflow}
\caption{Dataset construction workflow.
Left, image acquisition, reconstruction and three-dimensional component annotation.
Centre, projection of component polygons into selected images.
Right, manual review and correction, annotation export and station-level partitioning.
Dashed lines indicate projection rays; solid arrows indicate workflow progression.}\label{fig:workflow}
\end{figure}

Component outlines were delineated in reconstructed three-dimensional scenes and reprojected into selected images to reduce repeated boundary drawing across views. The projected annotations underwent image-by-image review and correction, followed by annotation export and station-level partitioning (Fig.~\ref{fig:workflow}).

\FloatBarrier

\topic{UAV image acquisition}
The study covered 22 PV stations in China: 18 in Ningbo and three in Shaoxing, Zhejiang Province, and one in Shenzhen, Guangdong Province.
The stations varied in roof and support height, component orientation and layout, and surrounding building context (Table~\ref{tab:sites}).

RGB images were acquired from January 2024 to January 2026 using a DJI Mavic 3E along oblique photogrammetric flight routes. Image dimensions were $5280\times3956$ pixels. Flight heights were 20~m relative to the roof in Ningbo and Shenzhen and 40~m in Shaoxing, with 80\% forward and 70\% side overlap.
The estimated nadir-equivalent GSD at these heights is approximately 0.5--1.1~cm/pixel; local image scale varies with viewing angle and roof geometry.
Station-specific acquisition months, reconstruction-input image counts and release counts are listed in Supplementary Table S7.

\begin{table}[!htbp]\centering\small
\caption{Study areas, flight heights and partition assignments.}\label{tab:sites}
\begin{tabularx}{\linewidth}{lrrX}\toprule
City & Stations & Height & Assignment\\\midrule
Ningbo & 18 & 20 m & 12 training stations; 6 independent test stations\\
Shaoxing & 3 & 40 m & Cross-city testing\\
Shenzhen & 1 & 20 m & Cross-city testing\\\bottomrule
\end{tabularx}\par\smallskip\raggedright Heights are relative to the roof.
\end{table}

\Needspace{6\baselineskip}
\topic{3D reconstruction and annotation}
Agisoft Metashape was used for aerial triangulation and bundle adjustment within a structure-from-motion photogrammetric workflow\cite{westoby2012}. Onboard real-time kinematic (RTK) positioning constrained camera locations without ground control points. Bundle adjustment yielded camera interior and exterior orientation parameters and lens distortion coefficients. The triangulation results and images were imported into iTwin Capture Modeler Center 2024 Update 1 (version 24.1.7.4760, Bentley Systems). Three-dimensional scene reconstruction retained the imported camera parameters without further optimization.

Three-dimensional component annotation was performed by a professional team, with different personnel responsible for annotation and quality checking.
In DasViewer V3.6.0.303Beta, each physical component was manually delineated along its outer boundary on the reconstructed three-dimensional model. Each resulting three-dimensional polygon received a distinct identifier. Adjacent components were labelled separately; internal cells and dividing lines did not define instance boundaries.
Annotations underwent three tiers of quality control: automated checks, manual review and expert spot checks.
Each three-dimensional polygon represented one physical component; its two-dimensional projections formed separate image-level observations.

\Needspace{5\baselineskip}
\topic{UAV image selection and reprojection}
To reduce redundant observations, image sparsification used the Reduce Overlap operation in Agisoft Metashape 2.1.3 after generating a three-dimensional scene surface model for coverage analysis. Surface coverage was set to 5, defining the target number of cameras observing a surface point from different angles. The retained images were used for undistortion, annotation reprojection and release.

Using the adjusted camera parameters exported from Metashape, a custom Python program undistorted the retained images with radial coefficients $k_1,k_2,k_3$ and tangential coefficients $p_1,p_2$.
Undistortion preserved image dimensions and principal-point coordinates without cropping; the corrected-image projection model used zero distortion coefficients.
Camera parameters were reformatted, with orientations expressed as $\varphi,\omega,\kappa$ angles.
The four corners of each three-dimensional component polygon were projected onto the image plane using the camera poses and corrected-image intrinsics.
The projected vertices were connected in their original order to form two-dimensional outlines.
The collinearity relationship was expressed in homogeneous coordinates as

\begin{equation}
\lambda_i\mathbf{p}_i=\mathbf{K}\mathbf{R}(\mathbf{P}_i-\mathbf{C}),\qquad i=1,2,3,4.
\label{eq:projection}
\end{equation}
Here, $\mathbf{P}_i=(X_i,Y_i,Z_i)^{\mathsf T}$ and $\mathbf{C}=(X_c,Y_c,Z_c)^{\mathsf T}$ denote the three-dimensional component corner and camera centre in the same world coordinate system. The image point is $\mathbf{p}_i=(u_i,v_i,1)^{\mathsf T}$. Matrix $\mathbf{R}$ rotates world coordinates into camera coordinates, $\mathbf{K}$ contains corrected-image intrinsics, and $\lambda_i$ is a projective scale factor.

The resulting two-dimensional polygons were geometrically intersected with the image extent and saved as image-associated shapefile (SHP) labels, retaining only their in-image portions. Subsequent annotations used pixel coordinates with an upper-left origin, $x$ increasing rightward and $y$ increasing downward.

\topic{Manual annotation correction}
Projected SHP labels were converted into per-image annotations for the ISAT image segmentation annotation tool and overlaid on the undistorted images. One author inspected all images and corrected their annotations, and a second author reviewed the complete corrected annotation set. Corrections addressed displaced outlines, missing components, erroneous projections, duplicate labels, and incorrectly merged or split instances. Fully occluded components were removed, whereas partially occluded components retained their complete projected outlines, including occluded regions.

Corrected objects were exported as COCO instance segmentation annotations\cite{r16}, with one record per object and clipped fragments retained within the same instance. Polygon vertices were preserved in \texttt{segmentation}, while \texttt{pycocotools} rasterization yielded foreground-pixel \texttt{area} and axis-aligned \texttt{bbox}, stored as $[x,y,\mathrm{width},\mathrm{height}]$. Conversion checked image--annotation associations, image dimensions, categories and coordinates, and assigned image and instance identifiers. Self-intersecting, degenerate or invalid polygons and empty rasterized masks were excluded and recorded. Instances with bounding-box width or height below five pixels were also excluded. Images confirmed to contain no targets remained in the COCO \texttt{images} list without instance records.

\topic{Station-level data partitioning}
To keep spatially dependent observations together\cite{roberts2017}, all images, annotations and derived crops from each station were assigned to the same partition. Twelve Ningbo stations were used for supervised training and six for within-city, cross-station testing; three Shaoxing stations and one Shenzhen station provided cross-city tests (Table~\ref{tab:sites}).

\topic{Release preparation}
Undistorted images were saved as JPEG without copying Global Positioning System (GPS), RTK or other source positioning metadata, preserving image dimensions and annotation coordinate alignment.

\section{Data Records}\label{sec:records}
\topic{Data access and organization}
PVIS-10K (version 1.0.0) is publicly available from Zenodo\cite{uavpvis2026} at \url{https://doi.org/10.5281/zenodo.23886365} as independently downloadable station-level ZIP archives.
Archives use \texttt{train/<station>} or \texttt{test/<station>} directories with three-digit station identifiers.
Each \texttt{images} folder contains full-resolution, undistorted JPEG images.
Each training and test station directory also contains COCO annotations in \texttt{instances.json}, a JavaScript Object Notation (JSON) file.

The comma-separated values (CSV) file \texttt{station\_splits.csv} records station identifiers (\texttt{station\_id}), cities (\texttt{city}) and training or test membership (\texttt{split}). The \texttt{evaluation\_role} field distinguishes training, within-city testing and cross-city testing. Relative paths to images and annotations for both partitions are recorded in \texttt{images\_path} and \texttt{annotations\_path}, respectively. README files describe dataset use, and \texttt{LICENSE.md} specifies the CC BY 4.0 licence (Table~\ref{tab:files}).

\begin{table}[!htbp]\centering\small
\caption{Core files in the dataset.}\label{tab:files}
\begin{tabularx}{\linewidth}{p{0.40\linewidth}l>{\raggedright\arraybackslash}X}\toprule
Path or file & Format & Content\\\midrule
\path{train/<station>/images/}\newline\path{test/<station>/images/} & JPEG & Full-resolution undistorted RGB images\\
\path{train/<station>/instances.json}\newline\path{test/<station>/instances.json} & COCO JSON & Training and test image records, component polygons and category definitions\\
\path{station_splits.csv} & CSV & Station, city, partition, role and relative paths\\
\path{README.md}\newline\path{README_zh-CN.md} & Markdown & Dataset description and usage instructions\\
\path{LICENSE.md} & Text & CC BY 4.0 licence notice\\\bottomrule
\end{tabularx}\end{table}

\topic{Image and annotation records}
Image filenames retain their acquisition basenames with the extension \texttt{.jpg}. Each station's COCO file contains \texttt{images}, \texttt{annotations} and \texttt{categories} records. Image records provide filenames and pixel dimensions; \texttt{annotations.image\_id} links each annotation to its image, and \texttt{category\_id} identifies the single category, \texttt{pv\_panel}. Image and annotation identifiers are local to each station file, with the same structure used for training and test annotations.

The \texttt{segmentation} field stores polygon vertices as $[x_1,y_1,x_2,y_2,\ldots]$ in pixel coordinates, with an upper-left origin, $x$ increasing rightward and $y$ increasing downward. The \texttt{bbox} field stores the top-left coordinates, width and height as $[x,y,\mathrm{width},\mathrm{height}]$ in pixels; \texttt{area} counts foreground pixels after rasterization. Multiple fragments of one instance are stored as separate vertex sequences within the same annotation.

Annotations identify per-image component observations, not persistent physical identities across views. Target-free images have image records but no associated instances. Partially occluded components retain projected outlines that include occluded portions.

\begingroup
\setlength{\intextsep}{6pt}
\setlength{\parskip}{2pt}
\section{Data Overview}\label{sec:overview}
Table~\ref{tab:counts} summarizes the composition of PVIS-10K by city and data partition.
The Ningbo training subset contains 5,266 images and 1,578,541 instance annotations, accounting for just over half of the images.
Among the test subsets, Shaoxing contains fewer images but more annotations than Ningbo, whereas Shenzhen is the smallest in both measures.

\begin{table}[H]\centering\small
\caption{Dataset size and partition composition.}\label{tab:counts}
\begingroup\setlength{\tabcolsep}{4pt}\renewcommand{\arraystretch}{1.04}
\begin{tabularx}{\linewidth}{@{}cl*{5}{>{\raggedleft\arraybackslash}X}@{}}\toprule
City & Split & Stations & Images & 2D instances & Physical & Instances\\
 & & & & & components & per image\\\midrule
\multirow[c]{2}{*}{Ningbo} & Training & 12 & 5,266 & 1,578,541 & 28,731 & 299.8\\
 & Test & 6 & 2,439 & 723,943 & 13,176 & 296.8\\
Shaoxing & Test & 3 & 1,906 & 1,096,688 & 18,091 & 575.4\\
Shenzhen & Test & 1 & 454 & 117,503 & 3,634 & 258.8\\
\midrule[0.3pt]
\multicolumn{2}{l}{\textbf{Total}} & 22 & 10,065 & 3,516,675 & 63,632 & 349.4\\\bottomrule
\end{tabularx}\endgroup\par\smallskip\raggedright Instances count per-image observations; physical components count distinct annotated objects.
\end{table}

Figure~\ref{fig:overview} shows station-level image counts, annotation counts and mean instances per image.
Station image counts range from 58 to 1,228, whereas mean instance counts range from 23.0 to 1,092.0 per image.
Station 013 supplies the most images, but stations 034 and 006 each average more than 1,000 instances per image.

\begin{figure}[!htbp]\centering
\includegraphics[width=\linewidth]{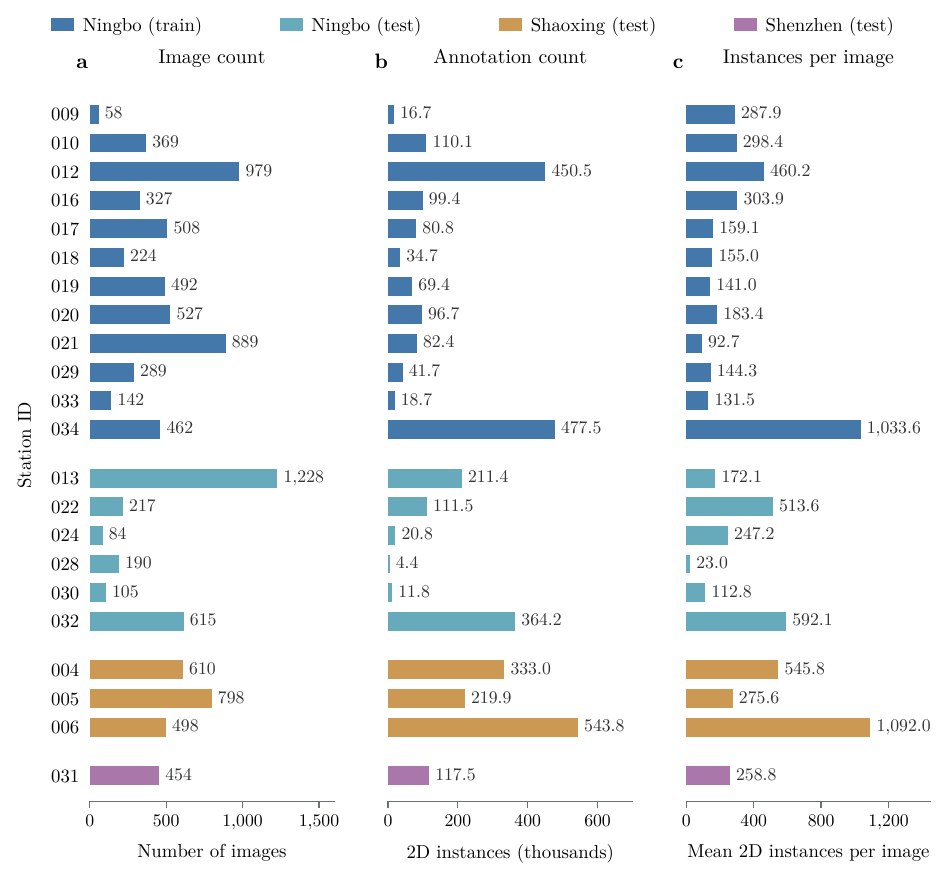}
\caption{Station-level dataset composition. \textbf{a}, number of images; \textbf{b}, two-dimensional instance annotations, in thousands; \textbf{c}, mean number of instances per image. Colours indicate the city and training or test subset.}\label{fig:overview}
\end{figure}
\FloatBarrier
\endgroup

\section{Technical Validation}\label{sec:validation}
Technical validation combined data quality checks with supervised and zero-shot model evaluation across all ten test stations (Table~\ref{tab:counts}).

\topic{Data quality checks}
An automated audit covered all 10,065 JPEG images and 3,516,675 instance annotations across the 22 stations.
Every image was fully decoded, and its dimensions were compared with the corresponding COCO record.
All released images were traced to the retained image subset by filename.
Checks covered file associations, identifier uniqueness within each station, category references, polygon validity and coordinate bounds.
All instance polygons were rasterized with \texttt{pycocotools} to independently recompute their mask areas and bounding boxes.

No missing or unreadable images, invalid references, duplicate identifiers, invalid polygon geometries or out-of-bounds coordinates were detected.
Recomputed mask areas and bounding boxes agreed with the stored values for every instance, and all bounding-box dimensions were at least five pixels.
The COCO image lists for the complete dataset retained 712 images without associated instances.
Station-level summaries included these target-free images when calculating instance density and described component pixel scale using bounding-box dimensions and the square root of mask area.

\topic{Model settings}
Evaluation included four supervised instance segmentation models and the text-prompted zero-shot Segment Anything Model~3 (SAM~3)\cite{r20}.
Both groups used the same station partitions, JPEG images and COCO reference annotations, with predictions evaluated at full-image resolution.
The supervised baselines were Mask R-CNN\cite{r5}, Hybrid Task Cascade (HTC)\cite{r17}, SOLOv2\cite{r18} and Mask2Former\cite{r6}, implemented in MMDetection\cite{mmdetection2019} 3.2.0.
HTC excluded the semantic segmentation branch; model-specific configurations are summarized in Table~\ref{tab:training}.

Supervised training used $512\times512$-pixel crops with a stride of 448 pixels.
Crops were excluded if at least 60\% of their channel elements were zero, and the training loader excluded crops without instances.
Labels were clipped to the same windows, removing instances with bounding-box width or height below five pixels.

Preliminary station-disjoint validation with Mask R-CNN within the 12 training stations used COCO instance segmentation average precision (AP) to select a ten-epoch schedule, which was applied to all four supervised models.
Each model was then reinitialized from the original pretrained weights and trained once on all 12 stations, using its epoch-ten checkpoint for testing.
Test stations were excluded from training and model selection.
Training used two NVIDIA A800 graphics processing units (GPUs), each with 80~GB memory.
Software environments, crop-boundary handling and the shared optimization schedule are detailed in Supplementary Methods.

\begin{table}[!htbp]\centering\small
\caption{Supervised instance segmentation training configurations.}\label{tab:training}
\setlength{\tabcolsep}{4pt}
\begin{tabularx}{\linewidth}{>{\raggedright\arraybackslash}p{0.25\linewidth}*{3}{>{\raggedright\arraybackslash}p{0.14\linewidth}}>{\raggedright\arraybackslash}X}\toprule
Parameter & Mask R-CNN & HTC & SOLOv2 & Mask2Former\\\midrule
Backbone & ResNet-50\cite{resnet2016} & ResNet-50 & ResNet-50 & ResNet-50\\
Feature processing & FPN\cite{fpn2017} & FPN & FPN & Multi-scale deformable-attention pixel decoder\\
Backbone initialization & Torchvision\newline pretrained & Torchvision\newline pretrained & Torchvision\newline pretrained & Torchvision\newline pretrained\\
Optimizer & SGD & SGD & SGD & AdamW\\
Base learning rate & 0.02 & 0.02 & 0.01 & 0.0001\\
Backbone learning-rate multiplier & 1 & 1 & 1 & 0.1\\
Weight decay & 0.0001 & 0.0001 & 0.0001 & 0.05\\
Momentum / $\beta$ & 0.9 & 0.9 & 0.9 & (0.9, 0.999)\\
Maximum gradient $L_2$ norm & -- & -- & 35 & 0.01\\
Batch size per GPU & 8 & 8 & 8 & 8\\
Global batch size & 16 & 16 & 16 & 16\\
Training epochs & 10 & 10 & 10 & 10\\\bottomrule
\end{tabularx}
\par\smallskip\raggedright FPN, feature pyramid network; SGD, stochastic gradient descent; AdamW, Adam with decoupled weight decay; --, no gradient clipping configured. Training used two GPUs without gradient accumulation, giving a global batch size of 16. Mask2Former normalization parameters and \texttt{query\_embed}, \texttt{query\_feat} and \texttt{level\_embed} parameters have no weight decay.
\end{table}

SAM~3 used Meta's official implementation and official pretrained weights downloaded from ModelScope, without training or fine-tuning on PVIS-10K.
The prompt ``individual solar panel'' was selected through preliminary comparisons on ten training images and fixed for all test images.

\topic{Full-image inference}
All five models used $512\times512$-pixel sliding windows with a stride of 448 pixels, retaining at most 100 candidates per crop.
Predictions were mapped to full-image coordinates, followed by cross-window edge-instance stitching using the criteria in Supplementary Table~S6.
Bounding-box non-maximum suppression retained higher-confidence predictions and suppressed lower-confidence predictions with an intersection over union (IoU) above 0.50 to reduce duplicates.
A subsequent mask-containment filter removed a smaller instance when at least 50\% of its area was covered by a larger retained instance, without comparing confidences.

\topic{Evaluation metrics}
Instance segmentation was evaluated using the COCO protocol\cite{r16}, reporting Mask AP over IoU thresholds from 0.50 to 0.95 in increments of 0.05. Mask AP50 and Mask AP75 denote AP at IoU thresholds of 0.50 and 0.75, respectively. Box AP and its corresponding threshold-specific metrics were reported as supplementary results. City-level AP was recalculated from all images, annotations and predictions within a city, rather than averaged across station AP values. Evaluation allowed up to 5,000 predictions per full image.

For pixel-level evaluation, predicted and ground-truth instance masks were separately united within each image.
Foreground IoU and the $F_1$ score were computed using their standard definitions\cite{reinke2024}.
City-level scores were calculated from confusion counts accumulated across all images in the city.

\topic{Supervised model evaluation}
HTC achieved the highest Mask AP among the four supervised models in each tested city (Table~\ref{tab:results}).
Mask AP was 72.72\% in Ningbo, 57.98\% in Shaoxing and 90.49\% at the single Shenzhen test station.
Performance on the cross-city Shenzhen set exceeded that on the within-city Ningbo set.

Station-level results revealed further variation within cities. HTC obtained 92.13\% Mask AP at Ningbo station 024, compared with 45.72\% at 013 and 45.17\% at 030. In Shaoxing, its Mask AP was 74.47\%, 78.08\% and 38.25\% at stations 004, 005 and 006, respectively. Supplementary Tables~S1--S4 provide complete station-level supervised results.

\begin{table}[!htbp]\centering\small
\caption{City-level instance segmentation and foreground coverage results (\%).}\label{tab:results}
\renewcommand{\arraystretch}{0.98}
\begin{tabular}{clrrrrr}\toprule
City & Model & Mask AP & AP50 & AP75 & IoU & $F_1$\\\midrule
\multirow[c]{5}{*}{Ningbo} & Mask R-CNN & 65.37 & 79.88 & 68.17 & 90.77 & 95.16\\
 & HTC & \textbf{72.72} & \textbf{85.87} & \textbf{75.57} & 90.42 & 94.97\\
 & SOLOv2 & 68.93 & 83.59 & 73.60 & 91.87 & 95.76\\
 & Mask2Former & 65.18 & 79.59 & 66.89 & \textbf{92.65} & \textbf{96.18}\\
 & SAM 3 & 4.75 & 7.68 & 4.95 & 89.49 & 94.45\\
\midrule[0.3pt]
\multirow[c]{5}{*}{Shaoxing} & Mask R-CNN & 55.30 & 68.75 & 60.86 & 90.14 & 94.81\\
 & HTC & \textbf{57.98} & \textbf{72.09} & \textbf{63.89} & \textbf{90.95} & \textbf{95.26}\\
 & SOLOv2 & 54.49 & 69.91 & 60.41 & 89.65 & 94.54\\
 & Mask2Former & 49.24 & 62.80 & 53.44 & 89.03 & 94.19\\
 & SAM 3 & 33.37 & 38.40 & 36.74 & 86.36 & 92.68\\
\midrule[0.3pt]
\multirow[c]{5}{*}{Shenzhen} & Mask R-CNN & 88.57 & 97.80 & 91.91 & 93.33 & 96.55\\
 & HTC & \textbf{90.49} & \textbf{98.91} & \textbf{94.15} & \textbf{94.80} & \textbf{97.33}\\
 & SOLOv2 & 88.54 & 97.90 & 94.03 & 94.58 & 97.21\\
 & Mask2Former & 87.42 & 98.71 & 89.85 & 92.86 & 96.30\\
 & SAM 3 & 2.46 & 3.77 & 2.30 & 90.28 & 94.89\\
\bottomrule\end{tabular}
\par\smallskip\raggedright Bold values indicate the highest score for each metric within each city.
AP50 and AP75 refer to mask-based evaluation. IoU and $F_1$ refer to the foreground union of instance masks. City-level AP is recomputed from pooled predictions; pixel metrics use accumulated confusion counts.
\end{table}

\topic{Zero-shot SAM~3 evaluation}
SAM~3 obtained Mask AP values of 4.75\%, 33.37\% and 2.46\% in Ningbo, Shaoxing and Shenzhen, respectively, below all corresponding supervised baselines. Its relatively higher instance performance in Shaoxing contrasted with the city-level pattern of the supervised models. Complete station-level results are provided in Supplementary Table~S5.

\topic{Analysis of evaluation results}
Instance matching and foreground coverage produced different model rankings (Table~\ref{tab:results}).
In Ningbo, HTC achieved the highest Mask AP, whereas Mask2Former achieved the highest foreground IoU of 92.65\%.
Despite its low Mask AP, SAM~3 retained foreground IoU values of 89.49\% in Ningbo and 90.28\% in Shenzhen.
Foreground unions discard instance identities, so duplicate, merged or split predictions can preserve coverage while reducing component-level matching accuracy.

\begin{figure}[!t]\centering
\includegraphics[width=\linewidth]{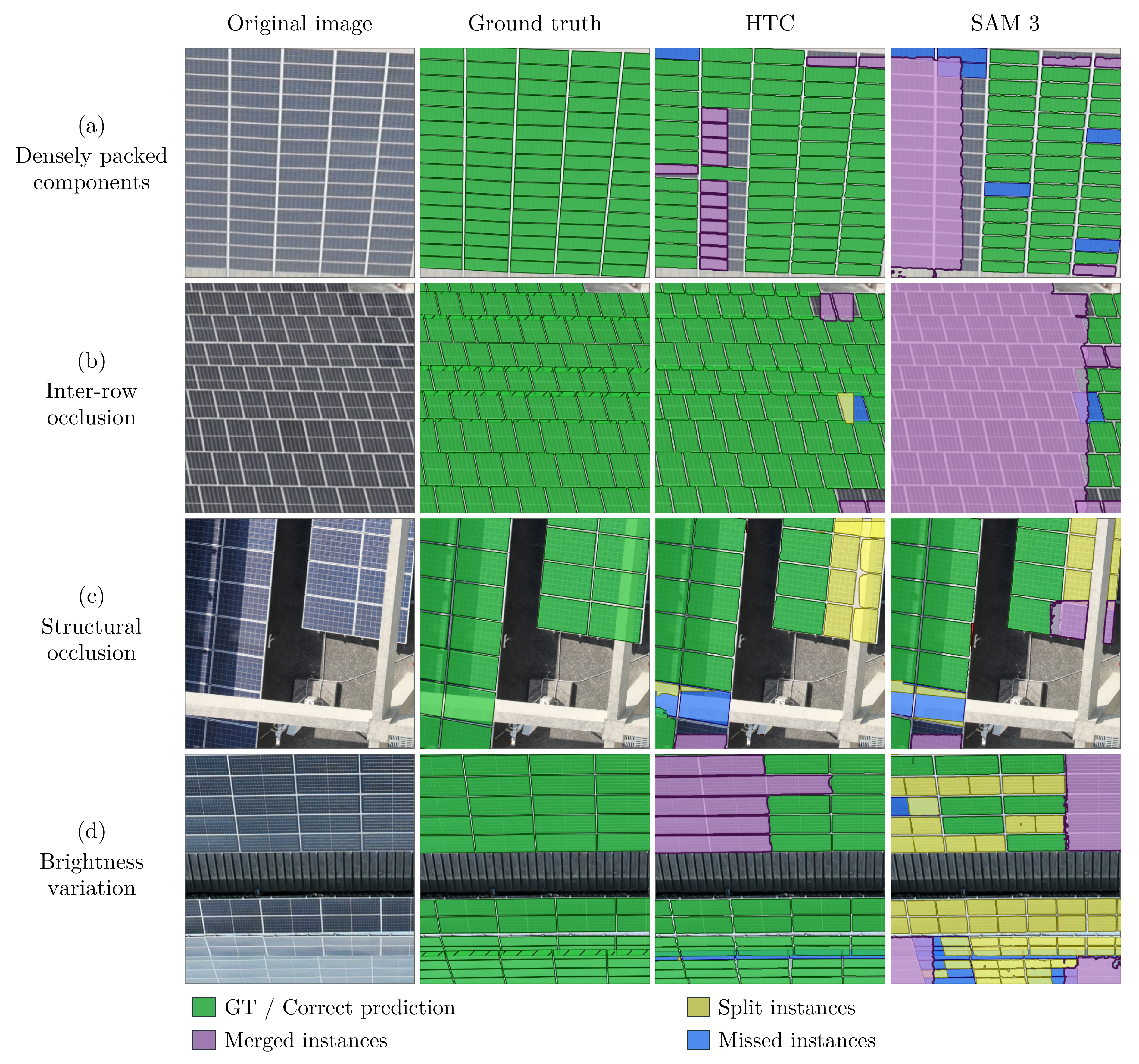}
\caption{Qualitative comparison of component-level instance segmentation. Columns show the original image, ground truth, HTC and SAM~3 predictions. Rows show \textbf{a}, densely packed components (station 022); \textbf{b}, inter-row occlusion (032); \textbf{c}, structural occlusion (028); and \textbf{d}, brightness variation (022). Crop and display scale are matched within each row. Green denotes ground-truth masks in the reference column and correct predictions in the model columns; yellow denotes split instances, where one PV component is divided into multiple predicted instances; purple denotes merged instances, where multiple PV components are combined into one predicted instance; and blue denotes missed instances, where a ground-truth component is not detected. Blue overlays mark missed ground-truth instances, not predicted masks. Colours illustrate the error types shown in these examples.}\label{fig:comparison}
\end{figure}

Figure~\ref{fig:comparison} compares HTC and SAM~3 predictions across four regions from three Ningbo test stations.
SAM~3 misidentifies boundaries between densely arranged components (Fig.~\ref{fig:comparison}a).
Under structural occlusion, predictions following visible portions differ from the full projected annotations (Fig.~\ref{fig:comparison}c).
Under inter-row occlusion, SAM~3 predicts instances spanning adjacent components (Fig.~\ref{fig:comparison}b).
In the region of contrasting brightness, SAM~3 incorrectly merges multiple PV components into a single predicted instance (Fig.~\ref{fig:comparison}d).

\begingroup\let\FloatBarrier\relax
\section{Usage Notes}\label{sec:usage}
\endgroup
Extract station archives into the directory structure described in Section~\ref{sec:records}, preserving image filenames and image--annotation associations. Combining station-level COCO files requires remapping image and instance identifiers and updating their links. Released images are already undistorted and require no further distortion correction. Annotation coordinates are pixel coordinates, not geographic coordinates.
Polygon annotations can be rasterized using \texttt{pycocotools}. Partially occluded components retain projected outlines that include occluded regions, rather than masks of visible surfaces alone.

Training crops can be generated with the repository's tools using the window size, stride and filtering rules specified under Model settings in Technical Validation and in Supplementary Methods.
Image and annotation crops must share window locations and edge handling.
Each station and its derived crops must remain in the same partition.
Any validation subset should be drawn from training stations, keeping stations disjoint between training and validation.

Baseline reproduction should follow the model settings reported in Technical Validation.
Predictions from other methods should be mapped to complete test images and evaluated against the released test annotations using the same evaluation protocol.

\begingroup\let\FloatBarrier\relax
\section*{Data Availability}
\endgroup
\addcontentsline{toc}{section}{Data Availability}
PVIS-10K (version 1.0.0) is openly available from Zenodo\cite{uavpvis2026} at \url{https://doi.org/10.5281/zenodo.23886365}.
The release includes undistorted JPEG images and COCO annotations for all 22 training and test stations, station-level partition definitions and supporting documentation.
These materials are licensed under the Creative Commons Attribution 4.0 International licence (CC BY 4.0; \url{https://creativecommons.org/licenses/by/4.0/}).
Three-dimensional models, camera parameters and intermediate annotation files are not included.

\section*{Code Availability}
\addcontentsline{toc}{section}{Code Availability}
Data-processing scripts, model configurations and inference and evaluation code are publicly available at \url{https://github.com/plzhang1998/PVDataset}.
The repository provides project-specific scripts and configurations; the MMDetection and SAM~3 frameworks must be installed separately.
Dependencies and execution instructions are documented in the repository.
Original project code and modifications are licensed under Apache License 2.0; third-party components retain their respective licences.
The trained checkpoints of the four supervised baselines are not distributed.

\phantomsection\addcontentsline{toc}{section}{References}
\setlength{\bibsep}{1.5pt}

\section*{Competing Interests}
\addcontentsline{toc}{section}{Competing Interests}
The authors declare no competing interests.

\section*{Funding}
\addcontentsline{toc}{section}{Funding}
This work was supported by the 2024 First Batch Yongjiang Talent Programme for Young Innovative Talents
(Grant No.~2024A-187-G), under the project
``Research and Application of High-transparency and High-brightness Object Target Detection and Visual Recognition Technology'', as well as by the National Natural Science Foundation of China (Grant No.~42371475).

\end{document}